\UseRawInputEncoding

\documentclass[a4paper, 10pt, conference]{ieeeconf}      

\IEEEoverridecommandlockouts                              

\title{\LARGE \bf
An Autonomous Driving Framework for Long-term Decision-making and Short-term Trajectory Planning on Frenet Space
}

\usepackage{cite}
\usepackage{amsmath,amssymb,amsfonts}
\usepackage{algorithmic}
\usepackage{graphicx}
\usepackage{textcomp}
\usepackage{xcolor}
\usepackage{subfig}
\usepackage{float}
\usepackage{multirow}
\usepackage{tikz}
\usepackage{hyperref}

\usepackage{amsthm}

\newcommand{\DrawBlackPercentageBar}[1]{%
  \begin{tikzpicture}
    \fill[color=black]   (0.0 , 0.0) rectangle (#1*8ex , 1.5ex );
    \fill[color=gray] (#1*8ex  , 0.0) rectangle (8.0ex, 1.5ex);
  \end{tikzpicture}%
}

\newcommand{\DrawRedPercentageBar}[1]{%
  \begin{tikzpicture}
    \fill[color=red]   (0.0 , 0.0) rectangle (#1*8ex , 1.5ex );
    \fill[color=gray] (#1*8ex  , 0.0) rectangle (8.0ex, 1.5ex);
  \end{tikzpicture}%
}

\def\BibTeX{{\rm B\kern-.05em{\sc i\kern-.025em b}\kern-.08em
    T\kern-.1667em\lower.7ex\hbox{E}\kern-.125emX}}

\author{Majid Moghadam$^{1}$ and Gabriel Hugh Elkaim$^{2}$
\thanks{$^{1}$Majid Moghadam is with Electrical and Computer Engineering Department,
        University of California, Santa Cruz, USA
        {\tt\small mamoghad@ucsc.edu}}%
\thanks{$^{2}$Gabriel H. Elkaim is with Electrical and Computer Engineering Department,
        University of California, Santa Cruz, USA
        {\tt\small elkaim@ucsc.edu}}%
\thanks{Submitted to International Conference on Robotics and Automation (ICRA 2021) for review}
}

\begin{document}

\maketitle
\thispagestyle{empty}
\pagestyle{empty}
\theoremstyle{remark}
\newtheorem*{remark}{\textbf{Remark}}

\begin{abstract}
In this paper, we present a hierarchical framework for decision-making and planning on highway driving tasks. We utilized intelligent driving models (IDM and MOBIL) to generate long-term decisions based on the traffic situation flowing around the ego. The decisions both maximize ego performance while respecting other vehicles' objectives. Short-term trajectory optimization is performed on the Frenet space to make the calculations invariant to the road's three-dimensional curvatures. A novel obstacle avoidance approach is introduced on the Frenet frame for the moving obstacles. The optimization explores the driving corridors to generate spatiotemporal polynomial trajectories to navigate through the traffic safely and obey the BP commands. The framework also introduces a heuristic supervisor that identifies unexpected situations and recalculates each module in case of a potential emergency. Experiments in CARLA simulation have shown the potential and the scalability of the framework in implementing various driving styles that match human behavior.
\\
\\
Code: \textnormal{https://github.com/MajidMoghadam2006/frenet-trajectory-planning-framework}
\end{abstract}

\section{INTRODUCTION}
In the past three decades, self-driving cars have witnessed considerable advancements in academic research and automotive development. Driving in highway traffic is a challenging task, even for a human, that requires intelligent decision-making for long-term goals and cautious short-term trajectory planning to execute decisions safely. 

Advanced driving assistance system (ADAS) is a hierarchical architecture that incorporates object detection, sensor fusion, planning, and control modules. Automakers and researchers leverage ADAS to approach the autonomous driving problem in a modular manner \cite{ziegler2014making}. Several methods have been proposed for the decision-making of autonomous vehicles on a highway driving task. Most of the studies approached as a control problem \cite{taylor1999comparative, hatipoglu2003automated}. Recently, deep reinforcement learning (RL) approaches have presented a decent alternative to the optimal lane-changing problem \cite{alizadeh2019automated, hoel2018automated}. However, none of the solutions provide a reliable method to translate the generated decisions to safe trajectories. 

Trajectory planning, on the other hand, has been addressed by multiple studies. Claussmann et al. \cite{claussmann2019review} distinguish the space configuration for the path planning into three main categories: i.e., sampling points \cite{li2014unified}, connected cells \cite{yu2016semantic}, and lattice representation \cite{werling2010optimal}. Sample-based decompositions normally provide a probabilistic method to randomly sample points from the feasible space and generate obstacle-free roadmaps. Although they are useful in local urban planning, the major drawback is that they do not guarantee that a solution will be found in a finite computation time, which would be disastrous in highway driving. Connected cells create an occupancy grid that is not memory efficient and introduce false indicative occupation with moving obstacles on the highway, making the approach a good option for decision-making but not for planning. On the contrary, lattice in motion planning provides a spatial structure of the path that includes the vehicle motion primitives. Lattice enables predictive planning based on the moving obstacles surrounding ego while considering the kinematic constraints, making this method a feasible representation for trajectory planning. In this work, we have utilized lattice representation to generate candidate trajectories and chose the optimal path among them. 
\begin{figure}[!t]
    \centering
    \includegraphics[width=\linewidth]{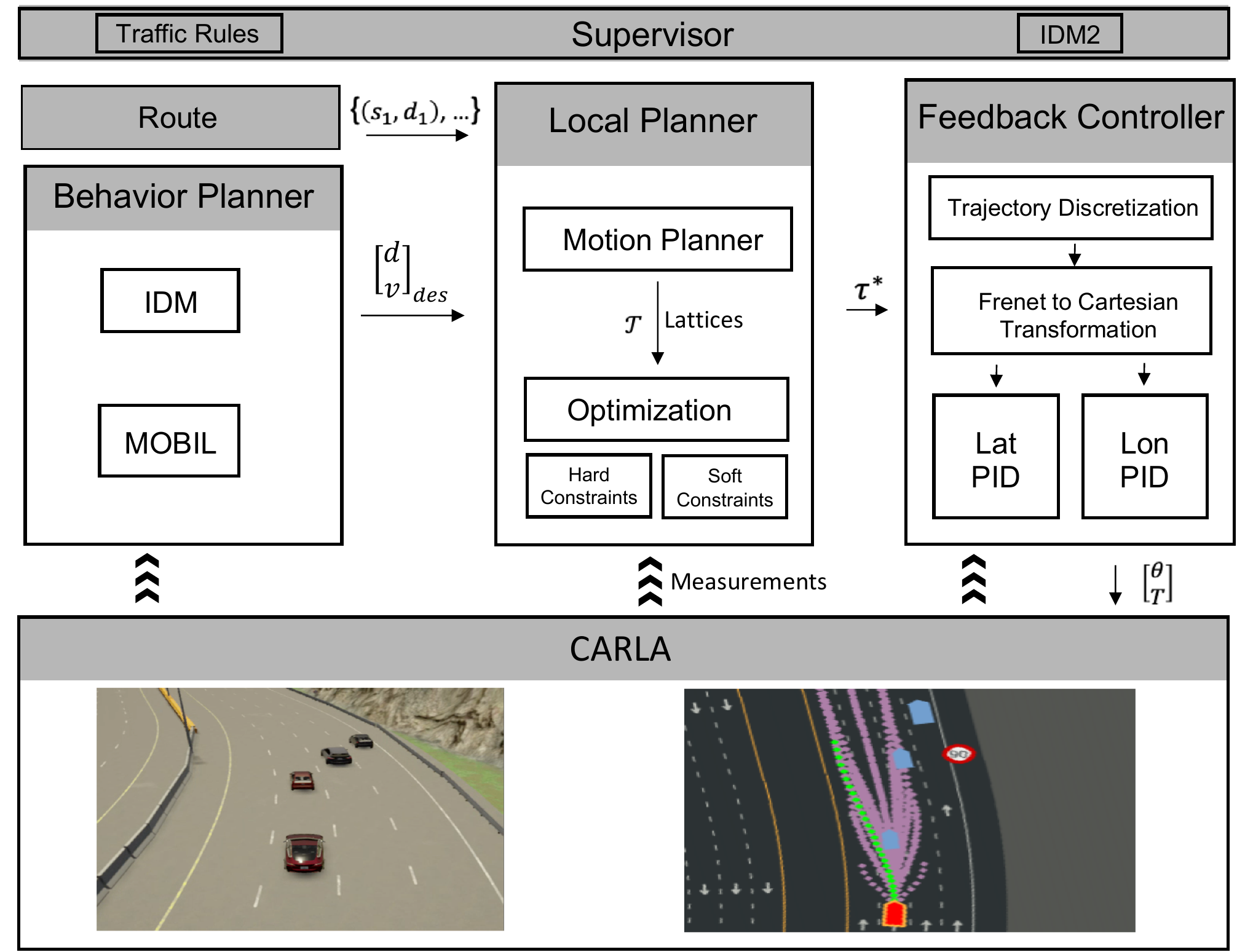}
    \caption{The proposed hierarchical architecture for the long-term decision-making and short-term trajectory planning in the Frenet space.}
    \label{fig:ADAS hierarchical arch.}
\end{figure}

The vast majority of the studies focused on generating collision-free trajectories using optimization techniques. For instance, in a recent distinguished work, Bertha-Benz \cite{ziegler2014trajectory} formalized the urban trajectory planning problem as a nonlinear optimization problem with constraint polygons for moving objects and used a Newton-type method to solve the optimal trajectory. Although Bertha's planning approach exhibited a promising outcome in urban driving, it may lack the intelligence and, as a result, lack safety on highway driving since the optimization attempts to find the short-term obstacle-free path, but it does not incorporate long-term goals. 

In this work, we have provided a rigorous framework for planning in autonomous driving in the highway driving task. Fig. \ref{fig:ADAS hierarchical arch.} summarizes the overall architecture that has been developed. The architecture addresses long-term decision-making based on the traffic situation to maximize ego driving performance and improve the traffic flow by respecting other drivers' projected decisions. The framework also provides a simple and scalable motion planning algorithm on the Frenet frame \cite{werling2010optimal} to generate safe and feasible polynomial trajectories to address short-term goals. We have introduced a novel obstacle avoidance method for velocity obstacles on the Frenet space that enables the motion planner to explore the driving corridors \cite{bender2015combinatorial} and generate spatiotemporal trajectories. Transferring the calculations to Frenet space makes the driving behavior invariant to the road curvatures and road slopes in three dimensions, which improves the optimization computations significantly and simplifies the cost function manipulation. The framework also includes a scalable supervisor module that controls the safety of the generated decisions and trajectories. The supervisor sends a recalculation command to the modules if an unpredicted situation appears during the path following. This significantly improves the safety and reliability of the algorithm. We have also shown the simplicity of configuring various driving styles using intuitive parameters from the framework that resemble human behavior. We have employed CARLA \cite{dosovitskiy2017carla} as a high-fidelity simulation that correctly reproduces the real-world vehicle dynamics and the city highway design and environment stochasticity.

\section{Framework Hierarchical Architecture}
Fig. \ref{fig:ADAS hierarchical arch.} summarizes the implemented hierarchical architecture in CARLA that incorporates the proposed framework for trajectory planning in the Frenet space. The behavior planner utilizes the sensory measurements and commands high-level actions to the local planner to produce a feasible and optimal trajectory. A feedback controller stabilizes the vehicle dynamics while tracking the commanded trajectory.  The framework also includes a supervisor where heuristic functions can be implemented to append multiple layers of safety and reliability, e.g. forward and side collision avoidance systems (CAS) and lane keeping assist (LKA).  In the following, we will elaborate on each layer individually.

\subsection{Behavior Planning}
\begin{figure} [!t]
    \centering
    \includegraphics[width=\linewidth]{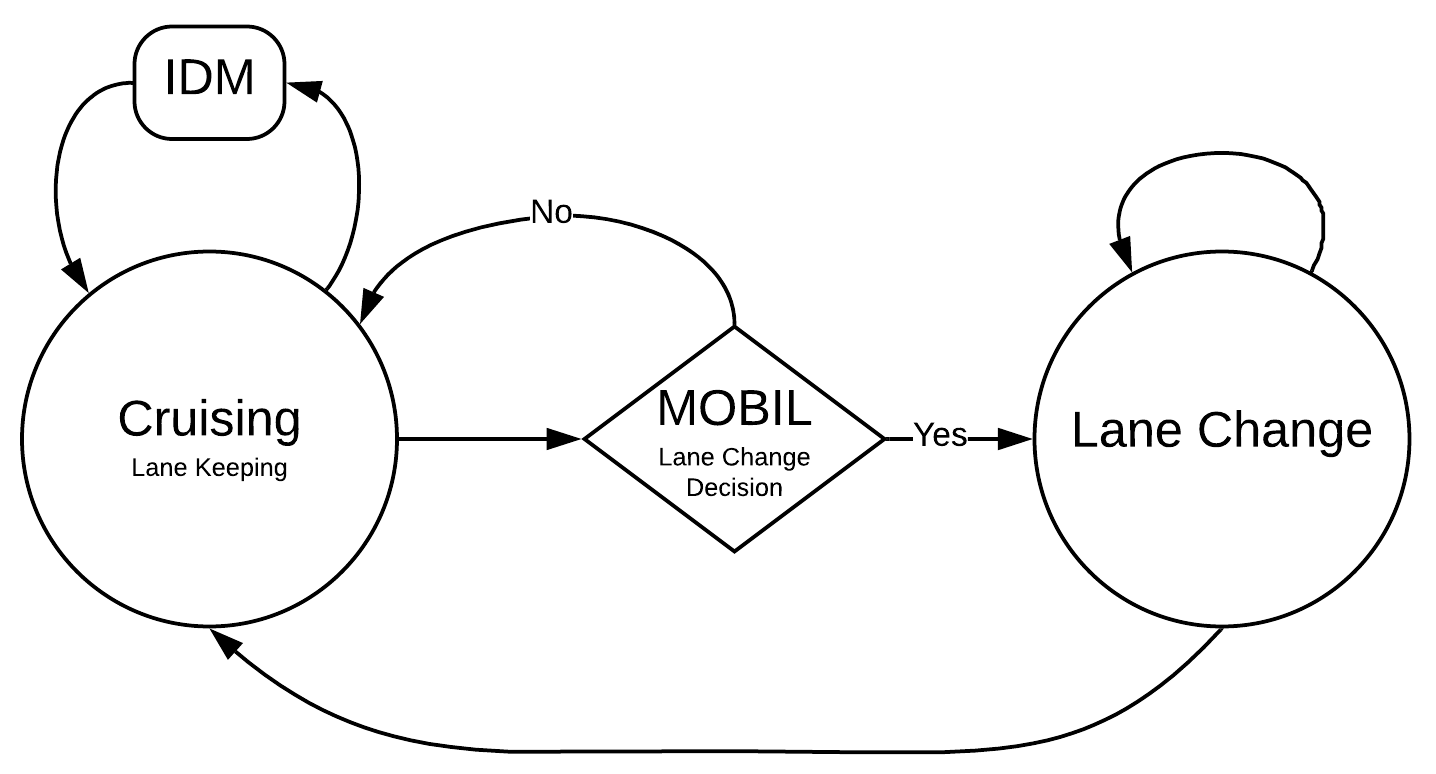}
    \caption{State transitions in the behavioral planning context}
    \label{f:FSM_IDM_MOBIL}
\end{figure}

Behavior planner (BP) is a key component in the planning architecture of autonomous driving. It generates a set of high-level driving actions or maneuvers to safely achieve desired driving missions under various driving constraints such as lane keeping, lane change, etc. Behavior planning module generates safe and efficient actions or maneuvers subject to the various constraints such as rules of the road, and the surrounding static and dynamic objects. We use Intelligent Driving Model (IDM) \cite{kesting2010enhanced} as an adaptive cruise control (ACC) and Minimizing Overall Braking Induced by Lane changes (MOBIL) \cite{kesting2007general} algorithms to cover all standard behavior decisions such as track speed, follow leader, decelerate to stop, stop, and lane change as illustrated in Fig. \ref{f:FSM_IDM_MOBIL}. 

As depicted in Fig. \ref{f:FSM_IDM_MOBIL}, the ego vehicle stays in the current lane with a desired speed computed by IDM module until MOBIL algorithm decides on a lane change. once lane change decision is made by MOBIL algorithm, the state of the ego transitions from cruising to lane change until lane change maneuver is done, then it continues to maintain its current lane in cruising mode.

\subsection{Local Planning}
\label{ss: local planning}
The decision-making layer in ADAS generates long-term decisions before sending them to the local planner (LP). The planner translates the commanded maneuver modes, such as LaneChangeRight and StayOnTheLane, along with the desired speed to optimal trajectories in a variable time-horizon window. The generated trajectories consist of two polynomials of time, concerning lateral and longitudinal movements. An optimization-based algorithm minimizes a user-defined cost function to generate an optimal trajectory. The heuristically manipulated cost function characterizes the commanded maneuver modes while capturing the optimal driving style, such as comfort and safety, and while constraining the trajectories to avoid dynamic obstacles surrounding the ego.
\newline
\subsubsection*{Frenet Frame}
\begin{figure}[!h]
    \centering
    \includegraphics[width=0.9\linewidth]{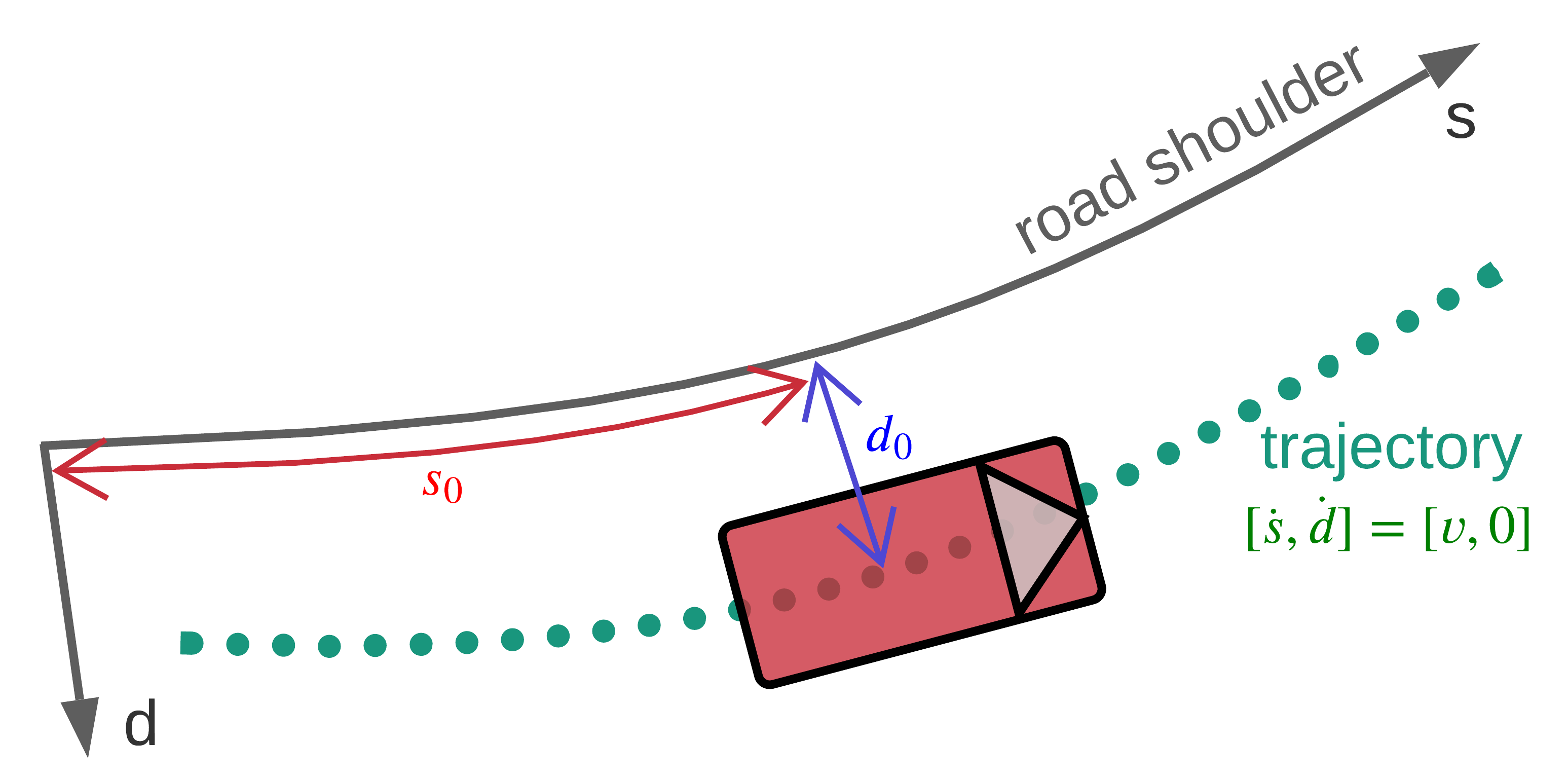}
    \caption{Frenet Frame visualization}
    \label{f:frenet frame}
\end{figure}

The driving behavior is less variant to the road curvatures than the surrounding actors' dynamic and interactions. Thus, it is more efficient to transform the calculations from the Cartesian coordinate to the Frenet frame\cite{werling2010optimal}, which specifies the vehicle's position in terms of longitudinal displacement along the road's arc ($s$) and the lateral offset from the road shoulder ($d$). Figure \ref{f:frenet frame}. illustrates the vehicle 2D position on the Frenet frame.
This representation is invariant to the road curvatures. i.e., a vehicle driving on the center of a lane with constant speed would have fixed $d$ and $\dot s$ values - dot indicates the time derivative throughout this paper. Following is the summarized procedure to transform from Cartesian to the Frenet frame,

\begin{itemize}
    \item Manually create a global route using a set of $n$ points, $\{(x_1,y_1,z_1), ..., (x_n,y_n,z_n)\}$, on the inertial frame and define
    \begin{equation}
       \bar x = \{x_1,...,x_n\}, \bar y = \{y_1,...,y_n\}, \bar z = \{z_1,...,z_n\}
    \end{equation}
    \item Define vector $\bar s$, with length $n$, where each element indicates the distance traveled since the first waypoint, i.e., the $i^{th}$ element is
    \begin{equation}
    s_i = s_{i-1} + \sqrt{(x_i - x_{i-1})^2 + (y_i - y_{i-1})^2 + (z_i - z_{i-1})^2}
    \end{equation}
    \item Interpolate three cubic spline curves \cite{maekawa2010curvature} $s_x(s)$, $s_y(s)$, and $s_z(s)$ for $(\bar s, \bar x)$, $(\bar s, \bar y)$, and $(\bar s, \bar z)$ pairs of vectors, respectively.
\end{itemize}

The forward transformation, shown in Fig. \ref{f:frenet frame}, from Frenet ($\mathcal{F}$) to Cartesian ($\mathcal{C}$) can be performed using the calculated splines, i.e,
\begin{equation}
(s_0,d_0)^\mathcal{C} = (x_0,y_0,z_0)
\end{equation}
where,
\begin{align}
    & x_0 = s_x(s_0) + d_0 \times \sin (s^{\prime}_x(s_0)) \nonumber \\
    & y_0 = s_y(s_0) + d_0 \times \sin (s^{\prime}_y(s_0)) \nonumber \\
    & z_0 = s_z(s_0) + d_0 \times \sin (s^{\prime}_z(s_0)) 
    \label{e: frenet to cart}
\end{align}
in which, the prime indicates the derivative w.r.t. variable $s$. There is no analytical solution for the inverse transform. We have utilized the approach introduced in reference \cite{fickenscher2018path} to estimate the Frenet states from Cartesian values by dividing the global path into smaller segments and locally approximating the divergence from the route.
\newline
\subsubsection*{Motion Planning in Frenet Frame}
\begin{figure}[!h]
	\centering
	\includegraphics[width=0.89\linewidth]{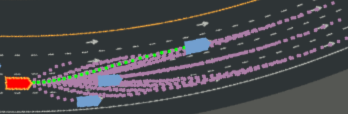}
	\caption{Lattices in Frenet frame for one time-step that align with the road curvature. The optimal trajectory is shown in green which takes the ego to the center line \cite{werling2010optimal}.}
	\label{f:lattices}
\end{figure}

Now that we defined the forward and inverse transformations, the planner can generate the desired trajectory in the Frenet format. The low-level controller receives the transformed optimal trajectory in the inertial frame and drives the vehicle along the path. At each time-step, the planner generates a set of candidate trajectories $\mathcal{T}=\{\tau_1, ..., \tau_m\}$ known as lattices that are shown in Fig. \ref{f:lattices}. 

The trajectories consist of $s(t)$ and $d(t)$, which are polynomials of time along Frenet axes
\begin{equation}
\begin{bmatrix}
s(t) & d(t)
\end{bmatrix} = 
\begin{bmatrix}
 \sum_{i=0}^{k} c_i t^i  & \sum_{i=0}^{r} a_i t^i
\end{bmatrix}
\end{equation}
where, $a$ and $c$ are the polynomial coefficients. The following remark enables us to calculate the polynomial coefficients. 
\begin{remark}
	For a generated path, 
	\begin{equation}
	    \tau_i = \Big\{(s_i(t), d_i(t)) \in \mathbb{R}^2 \hspace{1mm}\vline \hspace{1mm} t \in [t_0, t_f]=[0, T_i]\Big\}
	\end{equation}
	to be continuous w.r.t. the previous trajectory and to be dynamically feasible, the following conditions must hold
	\begin{align}
	& \begin{bmatrix} d_i(t_0) & \dot d_i(t_0) & \ddot d_i(t_0) & d_i(t_f) & \dot d_i(t_f) & \ddot d_i(t_f)\end{bmatrix} = \nonumber \\
	& \begin{bmatrix} d_{i-1}(T_{i-1}) & 0 & 0 & d_f & 0 & 0 \end{bmatrix}
	\end{align}
	\begin{align}
	& \begin{bmatrix} s_i(t_0) & \dot s_i(t_0) & \ddot s_i(t_0) & \dot s_i(t_f) & \ddot s_i(t_f)\end{bmatrix} = \nonumber \\
	& \begin{bmatrix} s_{i-1}(T_{i-1}) & 0 & 0 & v_f & 0 \end{bmatrix}
	\end{align}
	Here, we have defined $t_0 = 0$ as the initial time, $t_f = T_i$ as the time of arrival, $d_f$ as final lateral position, and $v_f$ as the vehicle velocity along $s$-axis at end of path. Note that we disregarded the lateral velocity at the beginning and the end of trajectories, so they align with the road's arc in both ends. Also, note that we defined six constraints for $d(t)$ and five for $s(t)$, which makes them form quintic ($r=5$) and quartic ($k=4$) polynomials, respectively. Since $t_0$ and $T_{i-1}$ are known values at each time-step, producing lattices boils down to identifying terminal manifolds: arrival time $t_f$, lateral position $d_f$, and speed $v_f$. The set $\mathcal{T}$ is produced by varying these three unknowns within the feasible ranges.
	\newline
\end{remark}
\subsubsection*{Numerical Optimization}
Since we generated the lattices, we can select the optimal trajectory $\tau^*$ from the set $\mathcal{T}$. Two kinds of constraints have been leveraged in this section. Hard constraints are utilized to eliminate infeasible trajectories that violate the vehicle's dynamical capabilities or potentially can make a collision. Soft constraints penalize the objective function in terms of safety, reliability, and comfort.

To evaluate the trajectories in terms of the hard constraints and generate the tracking reference for the feedback controller, we should generate higher-order information from each $\tau$, that is trajectories in Cartesian coordinate $\tau^{\mathcal{C}} = \{x(t), y(t), z(t)\}$, curvature $k(t)$, heading angle $\psi(t)$, velocity $v(t)$, acceleration $a(t)$, and jerk $j(t)$. To this end, for each $\tau$ we sample points from $s(t)$ and $d(t)$ polynomials with a constant sampling rate (CARLA's $dt$) to calculate a vector of samples for each variable. We use the following equations for the curvature and heading angle \cite{ziegler2014trajectory}
\begin{equation}
	k(t) = \frac{\dot{x}(t) \ddot{y}(t) - \dot{y}(t) \ddot{x}(t)}{\sqrt[3]{\dot{x}(t)^2 + \dot{y}(t)^2}}, \hspace{2mm}
	\psi(t) = \arctan (\frac{\dot{y}(t)}{\dot{x}(t)})
\end{equation}
Processing information in these vectors to check for hard constraint violations eliminates infeasible trajectories. To check for collision with dynamic obstacles, we must be able to anticipate the objects' future positions. Since the obstacles are moving vehicles with known states and dynamics, we can propagate the surrounding actors' positions up to the maximum time of horizon, $T_{max}$, in $\mathcal{T}$ and eliminate unsafe lattices that potentially collide with other obstacles,
\begin{equation}
    \mathcal{T} = \Big\{\tau \hspace{1mm} \vline \hspace{1mm} \tau \notin \mathcal{U} \Big\}
\end{equation}
where,
\begin{align}
\label{e: unsafe lattices}
    \mathcal{U} = \bigg\{ & \tau (s(t), d(t)) \hspace{1mm} \vline \hspace{1mm} 
    \Big( \exists \hspace{0.5mm} t^\prime \in [0, T]\Big)
    \Big( \exists \hspace{0.5mm} o(s_o(t), d_o(t)) \in \mathcal{O} \Big) \nonumber \\
    & \sqrt{\Big(s(t^\prime) - s_o(t^\prime)\Big)^2 + \Big(d(t^\prime) - d_o(t^\prime)\Big)^2} < r_c^2 \bigg\}
\end{align}
is the set of unsafe lattices that foreseeably collide with at least one obstacle from the obstacle set $\mathcal{O}$, with $r_c$ being the collision radius. Discovering the existence of $t^\prime$ in eq. \ref{e: unsafe lattices} between two objects is not a trivial problem, since it requires geometrical calculations. Assume that $\tau (s(t), d(t))$ is an arbitrary lattice in $\mathcal{T}$, and $\tau_o (s_o(t), d_o(t))$ is the obstacle's predicted trajectory. The problem is to find the existence of a $t^\prime$ at which the euclidean distance between $\tau$ and $\tau_o$ is less than the collision radius, $r_c$. Here, each trajectory forms a curved cylinder shape with base radius, $r_c$, where, we are checking if two shapes intersect in three-dimensional world. Two trajectories $\tau$ and $\tau_o$ intersect if
\begin{equation}
    \rho(t) = \sqrt{\Big(s(t) - s_o(t)\Big)^2 + \Big( d(t) - d_o(t)\Big)^2} - r_c
\end{equation}
has real roots. This can be discovered using Descarte's rule of signs, which indicates, a polynomial can have as many positive roots as it contains changes of sign, when the coefficients are ordered in terms of the exponents. Number of negative real roots can also be found by checking the $\rho(-t)$ polynomial. Repeating the same procedure for all pairs of lattices and obstacles eliminates the unsafe lattices.

\begin{figure}[!h]
    \centering
    \includegraphics[width=\linewidth]{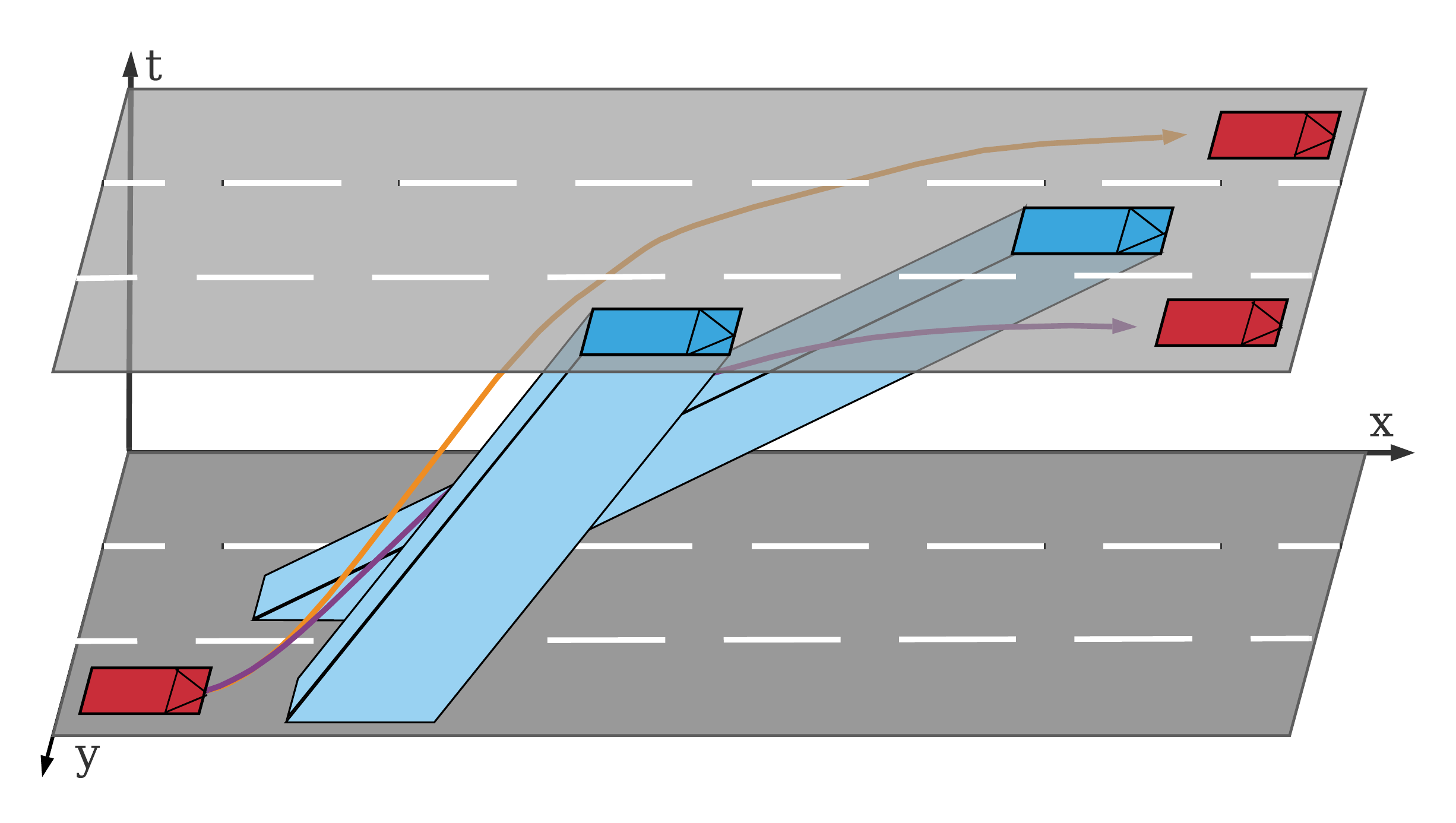}
    \caption{Spatiotemporal trajectories for moving obstacles and candidate paths visualized in driving corridors}
    \label{f:driving corridor}
\end{figure}

This process is basically exploring driving corridors - Fig. \ref{f:driving corridor} - to discover feasible lattices. Driving corridors incorporate the actors $x-y$ positions w.r.t. time. This enables us to find safe spatiotemporal trajectories that pass through the corridors as illustrated in Fig. \ref{f:driving corridor}. The remaining trajectory candidates are examined in terms of velocity
\begin{equation}
\begin{matrix}
v_{min} \leq ||v(t)|| \leq v_{max} & \forall t \in [0, T]
\end{matrix}
\end{equation}
and acceleration
\begin{equation}
\begin{matrix}
0 \leq ||a(t)|| \leq a_{max} & \forall t \in [0, T]
\end{matrix}
\end{equation}
that are safe, dynamically feasible, and legally allowed. The violating candidates get eliminated, which results in an updated set of candidate trajectories, $\mathcal{T}$.
When it comes to the soft constraints, it is possible to design an objective function that incorporates all of the required characteristics for the optimal trajectory, and utilize a numerical optimization to find the best trajectory among the candidates. The cost function to characterize the optimal trajectory $\tau^*$ is defined as
\begin{align}
\mathbf{J}(\tau) = w_{o} J_{o} + w_{v} J_{v} + w_{a} J_a + w_j  J_{j} + w_{\dot \psi} J_{\dot \psi}
\label{e: objective function}
\end{align}
where, $w_x$ weights identify the importance of each term. Tuning the coefficients is dependent to the vehicle dynamics and the trade-off between agility, safety, and comfort. In the following we discuss the individual terms in $\mathbf{J}$.
\begin{equation}
J_{o}(\tau) = (d(t) - d_{des})^2
\end{equation}
minimizes the vehicle's lateral offset to the target lane center. $d_{des}$ is a constant value, which, indicates the lateral position of the target lane on the Frenet coordinate. Although, MOBIL has already considered the safety before commanding $d_{des}$, here we append a second layer of safety for the Lane Change actions by incorporating the target lane safety in the cost function. Thus, it is possible for the LP to overlook the BP commands for the sake of safety and/or optimality.
\begin{equation}
J{v} = (v(t) - v_{des}(t))^2
\end{equation}
includes the vehicle speed error in the cost function to drive at the desired speed, which is generated by the BP layer. Similar to target lane, the speed commanded by the IDM can also be overwritten by the LP before being sent to low-level controller. Finally,
\begin{equation}
\begin{matrix}
J_a = a(t) ^ 2, 
& J_j = j(t) ^ 2,
& J_{\dot \psi} = \dot{\psi} (t) ^2
\end{matrix}
\end{equation}
suppress the vehicle acceleration, jerk, and yawing rate to improve the safety and comfort. 

Similar to Bertha \cite{ziegler2014trajectory}, it is possible to formulate the problem in convex optimization manner to find $\tau^*$ analytically. Although this approach is pragmatic and computationally efficient, the optimization is out of scope of this study. In addition, it is unclear how to incorporate the driving corridors in Bertha setup. Alternatively, we discretize $t_f, d_f,$ and $v_f$ within the feasible ranges and generate lattices $\mathcal{T}$ as shown in Fig. 
\ref{f:lattices}. Checking for hard constraints shrinks $\mathcal{T}$ to few trajectories in a crowded traffic. Finally, we utilize a simple linear search within $\mathcal{T}$ to find the optimal trajectory $\tau^*$ that minimizes $\mathbf{J}$ in eq. \ref{e: objective function}.

\subsection{Supervisor}
The supervisor is the highest layer in the hierarchy that can overwrite the output of each submodule or request for recalculation. The supervisor uses mathematical and heuristic functions and restrictions to append another safety layer to the system. Once BP commands the high-level decision, LP generates the trajectory, and the remaining layers work together to drive the ego on the path. To this end, BP and LP execute at lower frequencies than other layers. LP generates the optimal path based on the current situation of the surrounding vehicles. However, the surrounding human/autonomous drivers may perform unexpected maneuvers that jeopardize the safety of the generated trajectory. We employed a copy of the IDM controller (IDM2) with more conservative parameters to avoid forward collisions in the supervisor layer. At each time-step, IDM2 tracks the time to collision (TTC) with the leading and following vehicles in the target lane. If TTC violates the safety threshold, IDM2 raises the safety violation flag, and the supervisor calls the LP to recalculate the trajectory based on the current situation. In addition to this, we also implemented a simple heuristic function that checks the traffic rules violation of the highway maximum speed. This function can be enhanced by supplementing more traffic rules into the function.

\subsection{Low-level Controller}
It is possible to sample from the optimal trajectory $\tau^* = \{s^*(t), d^*(t))\}$ to generate a list of waypoints to track. The queue consists of $m=\frac{T}{dt}$ waypoints, where $T$  is the time of horizon of the path, and $dt$ is the simulation sampling rate. Since vehicle dynamics and the controllers are defined in the Cartesian frame, a Frenet to Cartesian transformation (eq. \ref{e: frenet to cart}) enables the controllers to calculate the desired throttle  ($T$) and the steering angle ($\theta$) values. A lateral PID model of the two-point visual controller \cite{salvucci2004two} is utilized to stabilize the steering angle. At each time-step, the controller pops the next two consecutive waypoints from the queue and calculates the desired steering angle. Since the waypoints are time labeled, the reference acceleration can be extracted from the list at each time-step. A longitudinal PID controller stabilizes the vehicle acceleration and speed by producing the desired throttle value.

\section{EXPERIMENTS}
\begin{figure}
    \centering
    \includegraphics[width=0.8\linewidth]{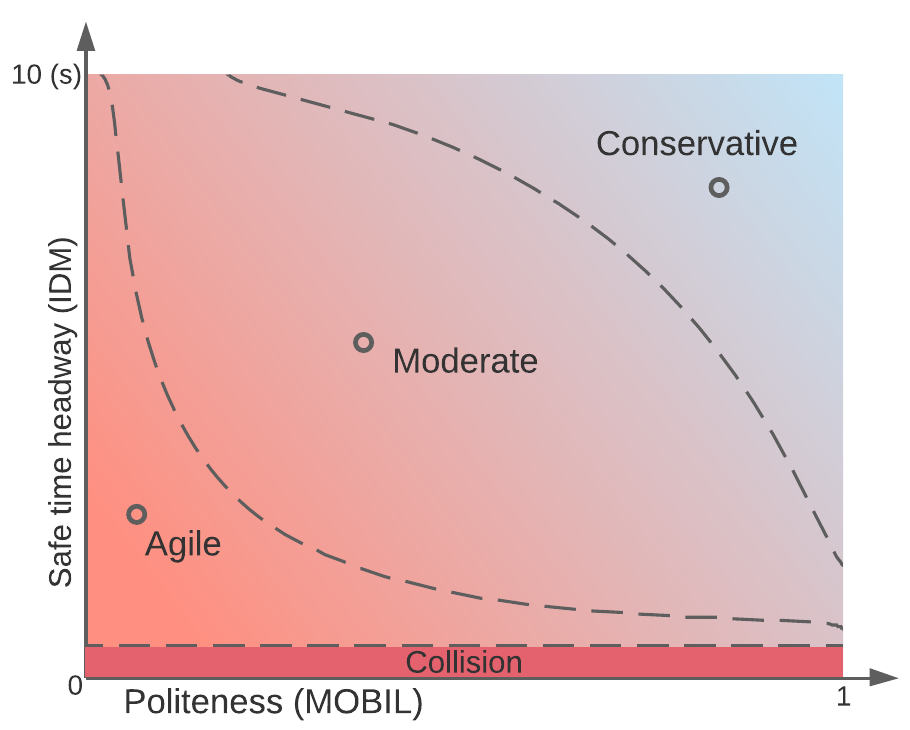}
    \caption{The Framework's trade-off between safety, agility, and the politeness among various driving styles}
    \label{fig:agility trade-off}
\end{figure}

This section provides a comprehensive evaluation and performance measurement for the proposed framework on CARLA v0.9.9.2 high-fidelity simulation environment. We selected TESLA model 3 as the ego dynamics and the highway loop available in TOWN04 to evaluate the proposed agents' performances. Trivially, the proposed agent's driving style is highly dependent on the parameter tuning of each layer of the hierarchy. This compromise introduces a trade-off between safety, comfort, and speed, i.e., fast agents tend to drive aggressively and vice versa. Fig. \ref{fig:agility trade-off} illustrates the trade-off and proposes a simple approach to achieve various driving styles by modifying only two parameters from IDM and MOBIL. Considering the measurements uncertainty, small safe-time-headway values can potentially cause accidents. The "Collision" region is provided to address this issue and prevent the user to chose driver parameter inside this region. The framework's simplicity and the proposed trade-off enabled us to introduce three different configurations: Agile, Moderate, and Conservative drivers. As the names suggest, each agent offers a different approach in the speed-maximization-lane-change-minimization trade-off. The remaining parameters are identical for all agents, however, the framework provides complete freedom in modifying parameters in all layers and designing various driving behaviors that match human characteristics. We utilized a common-sense method to tune the optimization soft constraint coefficients. Parameters that characterize hard constraints are selected according to the vehicle dynamics ($a_{max}$), highway regulations ($v_{max}$), and the safety criteria ($r_c$ and $v_{min}$).

\subsection{Qualitative Analyses}
This section provides a qualitative analysis based on the drivers' performance in several case studies. The scenarios cover situations where a self-driving car may commonly face in highway driving task, in addition, they target the intelligence, maneuverability, and the safety of the agents.

\subsubsection*{Case Study 1 (Intelligence-Safety)}
\begin{figure}[!t]
    \centering
    \includegraphics[width=\linewidth]{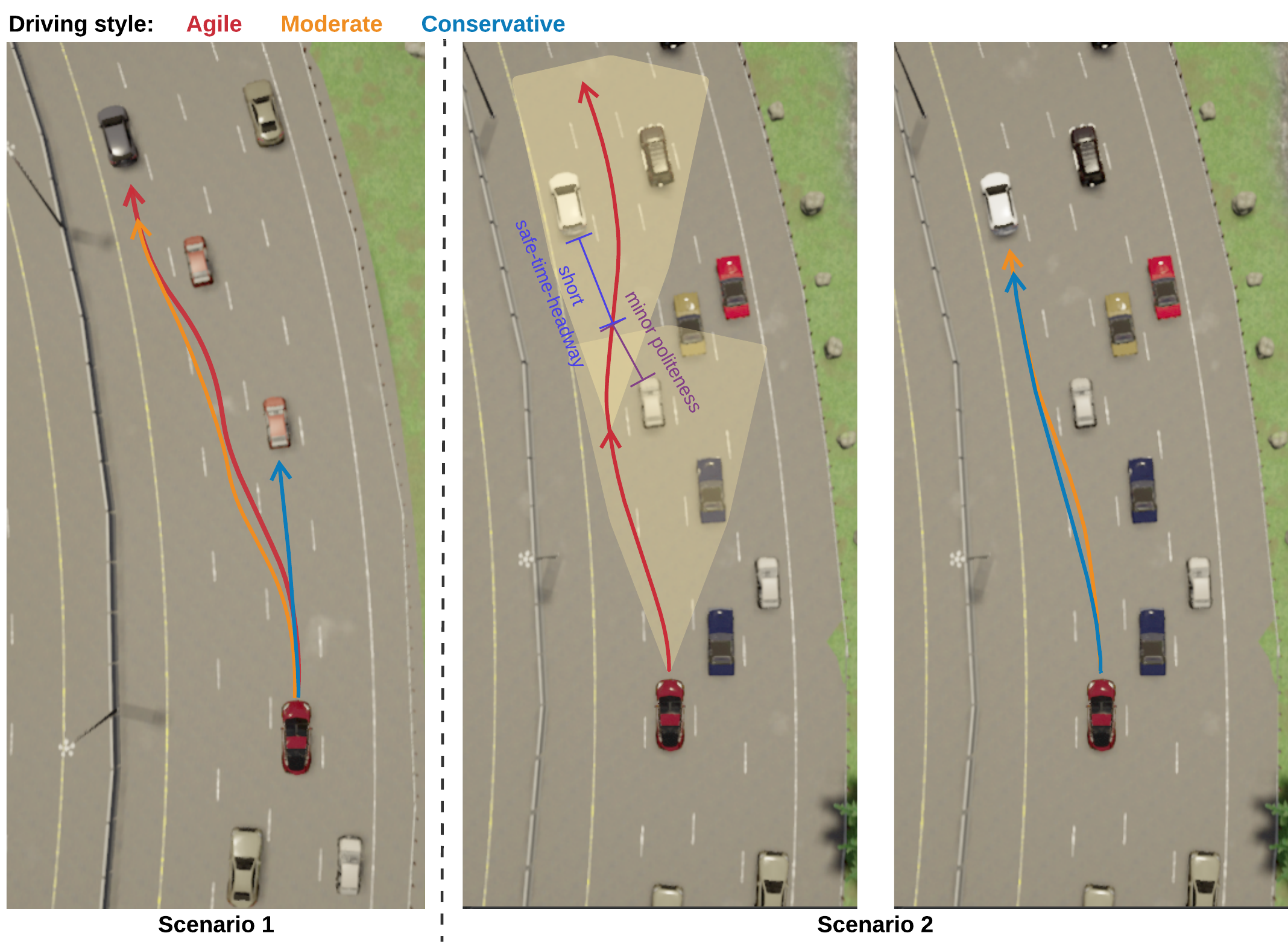}
    \caption{ Qualitative results for four proposed drivers in two case studies. Arrows show the agents tracked trajectories for two consecutive steps. Transparent regions show the convex shape of the generated lattices (recorded video of the agents' performances are submitted as supplementary files for review)}
    \label{fig:case studies}
\end{figure}

The scenario starts with a situation where the traffic flows on each lane at a different speed. The ego is driving on the second lane at 20 m/s. The lanes speed increase from right to left, making lane four the fastest passage. As illustrated in Fig. \ref{fig:case studies}, Agile and Moderate drivers make two consecutive lane changes to reach the fastest lane. However, the Conservative driver stays in the current lane until the left lane traffic becomes less sever then makes the lane change to left. Since lane three traffic is dense and moves faster than ego lane, a safe merging maneuver would take ego to the adjacent lane, however, the traffic in lane three would become temporarily slower. Agile and Moderate agents prefer the speed gain to politeness, making them slow down the traffic flow temporarily but finish the track much earlier than the Conservative agent. The scenario shows how the agents perform strategic and safe merging maneuvers to navigate through the traffic and eventually gain speed. This scenario also demonstrates the drivers' different approaches toward the agility-politeness trade-off, which shows the framework's compatibility to implement various driving styles by tuning only two parameters from the framework (Fig. \ref{fig:agility trade-off}). 

\subsubsection*{Case Study 2 (Maneuverability)}
Safely navigating through the cars to escape a traffic cluster requires situational awareness to perform a complex sequence of escaping maneuvers. In this scenario, the ego is tailgating a vehicle in lane three with 22 m/s. The traffic after this vehicle in lane three is smooth. A vehicle in lane four is moving slightly faster with 23 m/s. The traffic in lanes one and two are dense and slow because of an upcoming exit. Fig. \ref{fig:case studies} shows that Moderate and Conservative drivers move to the slightly faster lane on the left and keep driving there. Agile driver, on the other hand, performs an overtaking maneuver that assists the driver to escape the traffic and gain speed eventually. The MP lattices for two consecutive steps of agile agent are highlighted in yellow regions. While driving in lane four, most of the lattices that end up in lane three potentially violate the hard constraints (collision with moving obstacles). The remaining lattices keep a short safe-time-headway in IDM and a small politeness factor in MOBIL. Following those trajectories would require an aggressive maneuver, which may disrupt the traffic in lane three. The agile agent favors these lattices to trajectories that stay in lane four because of the driver's nature. This scenario confirms how the framework performs a complex sequence of maneuvers to achieve the desired driving style.

\subsubsection*{Quantitative Analysis}
In the qualitative analyses, we showed that the framework's situational awareness helped the drivers to generate complex trajectories and the maneuverability of the architecture enabled the agents to follow the generated trajectories. We also showed the qualitative differences between the driver's behavior in two case studies. Here we compare the driver's performance on randomly generated scenarios. The following quantitative metrics (percentages) have been used to compare the agents' performance on all of the scenarios,
\begin{itemize}
    \item Speed = $100(1 - \frac{\vert \text{average speed} - \text{target speed}\vert}{\text{max error}})$
    \item Comfort = $100 (1-w_c\frac{\text{average jerk}}{\text{max jerk}} - (1-w_c)\frac{\text{average yaw rate}}{\text{max yaw rate}})$
    \item Safety = $\frac{100}{n}\sum_{i=1}^{n} (1 - \frac{\text{min TTC}}{TTC_i})$
\end{itemize}
where $w_c$ is used to weigh the importance of factors and TTC stands for the frontal time-to-collision experiences. In the Safety equation we included $n$ steps of each scenario where ego tailgates a vehicle (TTC exists). For each scenario, the vehicles are spawned in random relative positions and target speeds to the ego. The scenarios start in an arbitrary position on the highway track, and the track length is 500 meters. Surrounding vehicles are randomly selected from a vehicle pool in CARLA, each one having different dynamics. We evaluated the agents' performance in 1000 scenarios and recorded the results in Table \ref{tab:random scenarios}. Overall, the agile driver showed a better approach toward gaining speed; however, it lacks safety and comfort. The safety issue becomes significant if the uncertainty level increases in the measurements. In contrast, the Conservative driver performed a more beneficial approach to safety and comfort but drove slow in most cases. The Moderate driver has displayed a satisfactory performance based on the provided metrics. This made the Moderate agent exhibit a better average percentage for all metrics in comparison with other drivers. Trivially, it is possible to migrate the Moderate driver point in Fig. \ref{fig:agility trade-off} to a desired sweet spot inside the region that matches the human driver's style. This demonstrates the flexibility of the proposed framework. 

\begin{table}
    \centering
      \begin{tabular}{c | c c c}
         & Conservative & Moderate & Agile\\
         \hline
         Speed
         & 53 \DrawBlackPercentageBar{0.53} & 74 \DrawBlackPercentageBar{0.74} & 81 \DrawRedPercentageBar{0.81}\\
         
        Safety
        & 60 \DrawRedPercentageBar{0.60} & 52 \DrawBlackPercentageBar{0.49}& 28 \DrawBlackPercentageBar{0.28}\\
        
        Comfort
        & 83 \DrawRedPercentageBar{0.83} & 75 \DrawBlackPercentageBar{0.70} & 47 \DrawBlackPercentageBar{0.47}\\
        \hline
        Average & 65 \DrawBlackPercentageBar{0.65} & 67 \DrawRedPercentageBar{0.67} & 52 \DrawBlackPercentageBar{0.52}\\
      \end{tabular}
    \caption{Performance metrics as percentages for three driving styles in 1000 randomly generated test scenarios}
    \label{tab:random scenarios}
\end{table}

\section{CONCLUSIONS}
In this paper, we introduced a hierarchical framework for decision-making and planning on highway driving tasks. IDM and MOBIL driving models have been utilized to maximize ego performance as well as the politeness wrt other drivers. A novel obstacle avoidance approach is introduced on the Frenet frame for the moving obstacles. The optimization explores the driving corridors to generate spatiotemporal polynomial trajectories to navigate through the traffic safely and obey the BP commands. The framework also introduced a heuristic supervisor that identifies unexpected situations and recalculates each module in case of a potential emergency.  Experiments  in  CARLA simulation  have  shown  the  promising performance  and  the  scalability  of  the framework  in  implementing  various  driving  styles  that  match human  behavior.

\bibliographystyle{IEEEtran}
\bibliography{references.bib}

\vspace{12pt}
\color{red}

\end{document}